\documentclass[runningheads]{llncs}
\usepackage[T1]{fontenc}
\usepackage{graphicx}
\usepackage{amsmath,amssymb,amsfonts,mathtools,bm}
\usepackage{booktabs}
\usepackage{multirow}
\newcommand{\argmin}{\operatorname*{arg\,min}}

\begin{document}
\title{CLEAR: Complex Learned Explicit Analytical Regularization for Ultra-Accelerated 4D Flow CMR Reconstruction}
\titlerunning{CLEAR: Complex Learned Explicit Analytical Regularization}
% If the paper title is too long for the running head, you can set
% an abbreviated paper title here
%
\author{German Shâma Wache$^{\dagger}$ \qquad
Sebastian Neumayer$^{\dagger}$% \and
%Third Author\inst{3}\orcidID{2222--3333-4444-5555}
}
\authorrunning{German Shâma Wache et al.}
% First names are abbreviated in the running head.
% If there are more than two authors, 'et al.' is used.
%
\institute{%Princeton University, Princeton NJ 08544, USA \and
%Springer Heidelberg, Tiergartenstr. 17, 69121 Heidelberg, Germany
%\email{lncs@springer.com}\\
%\url{http://www.springer.com/gp/computer-science/lncs} \and
${}^{\dagger}$Chemnitz University of Technology, Germany\\
\email{\{german-shama.wache,sebastian.neumayer\}@mathematik.tu-chemnitz.de}}
\maketitle              % typeset the header of the contribution
\begin{abstract}
While compressed-sensing regularizers enable interpretable reconstruction of 4D Flow CMR through transparent variational objectives, their hand-crafted nature is too restrictive under high acceleration.
State-of-the-art learning-based approaches mitigate this, but typically encode regularization implicitly through unrolled network modules, which limits their interpretability.
To address this limitation, we propose CLEAR, designed to combine the interpretability of compressed sensing with the flexibility of learned models.
To the best of our knowledge, it is the first learned regularizer for a 4D reconstruction task.
In the ultra-accelerated \(10\times\)--\(50\times\) regime of the CMRx4DFlow2026 challenge, CLEAR outperforms compressed sensing locally low-rank (LLR) and the popular variational network FlowVN, while using less than 10k parameters and preserving an interpretable regularization structure.

\end{abstract}

\section{Introduction}

Four-dimensional flow cardiovascular magnetic resonance (4D Flow CMR) provides non-invasive, time-resolved three-dimensional blood velocity fields, ultimately supporting the assessment of cardiovascular diseases \cite{markl2012time,dyverfeldt2015}.
Unlike conventional MRI, the object of interest is not only the complex image magnitude but also the blood velocity vector
\(  \mathbf v(\mathbf{r},t)=
    (\mathbf v_1(\mathbf{r},t),\mathbf v_2(\mathbf{r},t),\mathbf v_3(\mathbf{r},t))^\top
    \in \mathbb{R}^3,
\)
defined at spatial position \(\mathbf{r}\) and cardiac phase \(t \in \{1,\ldots, N_t\} \).
This velocity is typically encoded into the phase of \(N_e=4\) complex images \(\{\mathbf{x}_i\}_{i=0}^3\) according to
\begin{equation}
    \mathbf{x}_i(\mathbf{r},t)
    =
    \mathbf{x}_0(\mathbf{r},t)
    \exp\!\left(
        \jmath \pi
        \frac{\mathbf{v}_i(\mathbf{r},t)}{v_{\mathrm{enc}}}
    \right),
    \qquad i=0,1,2,3,
    \label{eq:pc_signal_model}
\end{equation}
where \(v_{\mathrm{enc}}\) is the velocity value corresponding to a phase of \(\pm \pi\), \(\mathbf{x}_0\) the flow-compensated reference image with $\mathbf{v}_0 = 0$, and \(\mathbf{x}_1,\mathbf{x}_2\), and \(\mathbf{x}_3\) encode the three velocity directions.
The velocity field is then obtained from the phase differences between the velocity-encoded images and the reference image.

The main limitation of 4D Flow CMR is its acquisition time \cite{bissell2023consensus}, since four velocity encodings must be acquired over multiple cardiac phases.
A common acceleration strategy is to acquire only a fraction of the k-space and to reconstruct the complex images before velocity extraction.
For a single velocity encoding \(i\in\{0,1,2,3\}\), let \(\mathbf{x}_i\in\mathbb{C}^{N_t\times N_x\times N_y\times N_z}\) denote the discretized image of interest on a $ N_x\times N_y\times N_z$ spatial grid across $N_t$ time frames.
Further, we require the spatial Fourier transform \(\mathcal F\)  acting time-frame-wise, the scanner $N_c$-channel coil sensitivity operator \(\mathbf S\),  the undersampling mask \(\mathbf M_R\) corresponding to an acceleration factor of $R$, and the measurement noise \(\boldsymbol{\eta}\).  
Assuming Cartesian sampling, the corresponding accelerated k-space measurement \(\mathbf{y}_i\in\mathbb{C}^{N_t\times N_c\times N_x\times N_y\times N_z}\)
obeys
\begin{equation}
    \mathbf{y}_i
    =
    \mathbf{M}_R\mathcal{F}\mathbf{S}\mathbf{x}_i
    +
    \boldsymbol{\eta}.
    \label{eq:forward_model}
\end{equation} 

Reconstructing \(\mathbf{x}_i\) from \(\mathbf{y}_i\) is ill-posed since only a fraction \(1/R\) of the k-space is acquired. The popular variational approach addresses this inverse problem by minimizing an objective consisting of a data-fidelity term and a carefully chosen regularizer \(\mathcal R\) (promoting desired properties of $\mathbf{x_i}$), namely
\begin{equation}
\label{eq:optimization}
\widehat{\mathbf x}_i
=
\argmin_{\mathbf x}
\frac{1}{2}
\left\|
    \mathbf{M}_R\mathcal{F}\mathbf{S} \mathbf x - \mathbf y_i
\right\|_2^2
+
\mathcal{R}(\mathbf x).
\end{equation}
Classical choices are compressed sensing (CS) regularizers, such as total variation (TV) \cite{rudin1992tv} or the locally low-rank (LLR) \cite{trzasko2011llr}.
However, their hand-crafted structure becomes restrictive at very high acceleration.
Recent CMR reconstruction methods improve expressiveness by complex learned architectures.
This includes region-focused transformer GAN reconstruction \cite{lyu2023regionfocused}, personalized federated reconstruction \cite{lyu2023federated}, and unrolled low-rank and sparse  networks~\cite{wang2023odls}. 
Dynamic settings additionally incorporate recurrent models that exploit temporal dependencies~\cite{qin2019crnn},  spatio-temporal attention \cite{lyu2024stadnet}, and deep separable spatio-temporal learning~\cite{wang2025deepseparable}.
More recent advances  include Mamba-based global modeling with uncertainty estimation~\cite{huang2025mambamir}, physics-informed synthetic data for improved generalization~\cite{wang2025pisf}, and diffusion-prior implicit neural representations~\cite{lyu2026dpinr}. 
For 4D Flow CMR, FlowVN \cite{vishnevskiy2020flowvn} introduced an unrolled variational network with learned spatio-temporal filters, while FlowMRI-Net \cite{jacobs2025flowmrinet} proposed a self-supervised unrolled network with complex recurrent modules. These methods are expressive, but their priors are typically implicit in network modules, limiting interpretability.
To compromise between interpretability and expressivity, a middle-ground route consisting of learning an explicit regularizer for \eqref{eq:optimization} emerged in 2D inverse problems; see  \cite{hertrich2026learning} for an overview.
A notable example is the learnable fields-of-experts regularizer \cite{roth2009fields}, which was repopularized in \cite{goujon2023convex,goujon2024weakly} with a recent extension to 3D in \cite{wache2026wcrr}.
To the best of our knowledge, this route has not yet been explored for 4D inverse problems.
Our contributions are thus the following:
\begin{itemize}
    \item We propose a learned regularizer with analytical gradient, enabling complex-valued 4D Flow CMR reconstruction based on \eqref{eq:optimization}.

    \item We condition the regularizer on the acceleration factor, allowing it to handle a wide and continuous range of undersampling rates.

    \item We evaluate the method in an ultra-accelerated \(10\times\)--\(50\times\) regime and compare it against CS and deep learning approaches.
\end{itemize}

\section{Method}

\subsection{Regularizer architecture}

Following the multi-subspace strategy in \cite{vishnevskiy2020flowvn}, the proposed regularizer replaces costly 4D spatio-temporal regularization with four branches operating in complementary 3D subspaces.
Specifically, let \(\mathbf x\in\mathbb C^{N_t\times N_x\times N_y\times N_z}\) be a 4D image.
We order its axes as $t, x, y, z$ and write $\prec$ for this order. 
For an axis $a \in \mathcal{A} \coloneqq \{t, x, y, z\}$ and a slice index $b \in \{1, \dots, N_a\}$, let $a_1 \prec a_2 \prec a_3$ be the elements of $\mathcal{A} \setminus \{a\}$ and denote by $P_{a,b} \colon \mathbb{C}^{N_t \times N_x \times N_y \times N_z} \to \mathbb{C}^{N_{a_1} \times N_{a_2} \times N_{a_3}}$ the linear extraction operator that selects the $b$-th 3D hypersurface orthogonal to $a$.
Then, we define the 4D regularizer
\begin{equation}
\mathcal{R}^{\text{4D}}(\mathbf{x}) = \frac{\lambda}{4} \sum_{a\in\mathcal A} \sum_{b=1}^{N_a} \mathcal{R}_{a^\perp}^{\text{3D}}\big( P_{a,b}\mathbf{x} \big), \quad \lambda > 0,
\label{eq:regularizer_4d}
\end{equation}
$\mathcal{R}_{a^\perp}^{\text{3D}}$ being the regularizer branch acting in the 3D subspace ${a^\perp}$ orthogonal to axis $a$, and whose parametrization is given below.

\paragraph{Regularizer branch parametrization.}
We parametrize the regularizer branch acting in the 3D subspace ${a^\perp}$, $a \in \mathcal{A}$, with the fields-of-experts formulation

\begin{equation}
    \mathcal R_{a^\perp}^{\text{3D}}(\mathbf z)
    =
    \sum_{c=1}^{C}
    \sum_{n=1}^{N_{a_1}}
    \sum_{n'=1}^{N_{a_2}}
    \sum_{n''=1}^{N_{a_3}}
    \psi_{{a},c}
    \bigl(
        \left|(W_{a} \mathbf z)_{c,n,n',n''}\right|_{\varepsilon_0}
    \bigr),
    \label{eq:regularizer_3d_branch}
\end{equation}
where \(W_{a}=[W_{{a},1},\ldots,W_{{a},C}]^\top\) is a learnable complex convolutional operator with \(C\) output channels applied to \(\mathbf z\in\mathbb C^{N_{a_1}\times N_{a_2}\times N_{a_3}}\), \(|\cdot|_{\varepsilon_0} \coloneqq \sqrt{|\cdot|^2+\varepsilon_0^2}\) with \(\varepsilon_0 = 10^{-6}\) is applied component-wise to the complex filter responses, and the potential \(\psi_{{a},c}\) acts on the magnitude response of channel \(c\).
Following \cite{hertrich2026learning,goujon2024weakly}, we build the potentials $\psi_{a,c}$ from the smoothed $\ell_1$ family $s_{\beta} = \frac{\beta}{2}[\vert\cdot\vert^{2}-(\vert\cdot\vert-\beta^{-1})_{+}^{2}]$ with $\beta>0$:
For each axis $a$, we use the $1$-weakly convex base potential $\phi_{\beta_{a}} = s_{\beta_{a}} - s_{1}$, rescaled per output channel as $\psi_{a,c} = \alpha_{a,c}^{-2}\phi_{\beta_{a}}(\alpha_{a,c}\,\cdot)$, where $\beta_{a}>1$ and $\alpha_{a,c}>0$ are learnable.

\begin{remark} \label{remar_1}
The channel-wise rescaling $\psi_{a,c}$ preserves the 1-weak-convexity of the base potential. Hence, with spectrally normalized convolutional operators ($\|W_{a}\| = 1$, $a \in \mathcal A$), each \(\mathcal R_{a^\perp}^{\text{3D}}\) is \(1\)-weakly convex, making the full regularizer \(\mathcal R^{\text{4D}}\) in \eqref{eq:regularizer_4d} \(\lambda\)-weakly convex.
The learnable regularizer is therefore allowed to be nonconvex and thus more powerful than purely convex penalties, while
its negative curvature remains bounded via the learnable parameter $\lambda$, potentially also leading to more stable minimization of the objective in \eqref{eq:optimization}.
\end{remark}

\begin{remark}
The smooth magnitude and smooth potentials make each regularizer branch $\mathcal R_{a^\perp}^{\text{3D}}$ differentiable, with analytical gradient
\(
    \nabla\mathcal R_{a^\perp}^{\text{3D}}(\mathbf z)
    =
    W_{a}^\ast
    (\psi_{{a},c}\circ|\cdot|_{\varepsilon_0})'
    (W_{a}\mathbf z),
\)
where the derivative acts component-wise on the complex response.
Hence, $\nabla \mathcal{R}^{\text{4D}}(\mathbf{x}) = \frac{\lambda}{4} \sum_{a \in \mathcal A} \sum_{b=1}^{N_a} P_{a,b}^\ast \nabla \mathcal{R}_{a^\perp}^{\text{3D}}( P_{a,b}\mathbf{x})$, enabling gradient-based optimization of the objective in \eqref{eq:optimization}.
\end{remark}

A visual illustration of the regularizer architecture can be seen in Figure \ref{fig:reg}. In practice, each convolutional operator \(W_{a}\) is normalized as required by Remark~\ref{remar_1} and implemented as a cascade of three bias-free, unit-stride, unit-padding complex convolutions with \(3\times 3\times 3\) kernels, giving an effective receptive field of \(7\times 7\times 7\). The cascade uses \(2\), \(4\), and \(C=16\) output channels, respectively. The first layer of the cascade is constrained to have zero-mean filters, so that the regularizer responds to local variations rather than global intensity offsets, in the spirit of classical sparsifying filters such as finite differences and wavelets.

\paragraph{Acceleration-conditioned regularizer.}
We condition the regularizer $\mathcal R^{\text{4D}}$ on the acceleration factor $R$ by using the modified potential scales 
\begin{equation} \label{eq:conditioning}
\alpha_{{a},c}(R) = R\exp\bigl(s_{c_{{a},c}}(1/R)\bigr) \text{ for }
R\in[R_{\min},R_{\max}],
\end{equation}
where \smash{\(s_{c_{{a},c}}\)} is a linear spline with \(K\) equidistant knots on \([1/R_{\max},1/R_{\min}]\) and learnable parameters \(c_{{a},c}\).
This enables $\mathcal R^{\text{4D}}$ to handle a wide continuous range of undersampling rates and its effectiveness is demonstrated in Section \ref{sec:Results}.
In practice, we used $R_{\min} = 9.0$, $R_{\max} = 51.0$ and $K=5$.

\subsection{Training and inference}

Our goal is to learn a configuration of $\mathcal R^{\text{4D}}$ that yields high-quality reconstructions through the variational formulation \eqref{eq:optimization} across the entire range of $R$.
We perform the underlying minimization with the \emph{non-monotone accelerated proximal gradient (nmAPG)} algorithm \cite[Supp.\ Thm.\ 4]{li2015apg}, guaranteeing convergence to a critical point.
The algorithm is initialized with the zero-filled reconstruction and terminated once the relative change between iterates falls below $10^{-2}$.

We train this reconstruction model end-to-end using a supervised $\ell_2$ loss.
This gives rise to a bilevel optimization problem:
the reconstruction itself constitutes the lower-level problem \eqref{eq:optimization}, while the upper-level problem adjusts the configuration of $\mathcal R^{4D}$ to minimize the reconstruction loss.
A central computational challenge is differentiating through the solution of \eqref{eq:optimization}.
To make this backward pass tractable, we employ Jacobian-free backpropagation \cite{fung2022jfb,bolte2023onestep}.
We refer to \cite{hertrich2026learning} for details on the bilevel learning of (generic) regularization functionals.

At inference, given a four-point velocity encoded accelerated 4D Flow CMR acquisition, the reference image $\mathbf x_0$ and the velocity-encoded images $\mathbf x_1$, $\mathbf x_2$ and $\mathbf x_3$ are all reconstructed independently.
Then, the velocity field is recovered as  $\mathbf v_i = \frac{v_{\mathrm{enc}}}{\pi}
    \arg\!\left(
        \mathbf x_i \overline{\mathbf x_0}
    \right)$, $i=1,2,3$.

\begin{remark} We refer to the proposed method as CLEAR: a \textbf{C}omplex,
\textbf{L}earned, \textbf{E}xplicit, \textbf{A}nalytically differentiable
\textbf{R}egularizer. The name also reflects the method's central idea: an interpretable learned regularizer (see Section~\ref{sec:interpretability}) is inserted into an explicit/clear variational objective to be minimized.
\end{remark}

\section{Experiments and Results}\label{sec:Results}

\subsection{Dataset and experimental setup}
We use the CMRx4DFlow2026 challenge data from the CMRx reconstruction series, whose previous datasets and challenge summaries are described in \cite{wang2024cmrxrecon,wang2025cmrxrecon2024,wang2025foundation,lyu2025cmrxreconchallenge,wang2025cmrxmotion,wang2026cmrxrecon2024summary}.
The full collection contains over 400 multi-center, multi-vendor Cartesian 4D Flow CMR cases, including more than 300 aortic cases, acquired from more than 10 centers, 4 vendors (GE, Philips, Siemens, United Imaging), 3 field strengths (1.5T, 3T, 5T), and over 6 anatomical regions.
The reported acquisition range is \(200\times200\times40\) to \(450\times450\times90~\mathrm{mm}^3\) FOV, \(0.9\)--\(3.0~\mathrm{mm}\) spatial resolution, \(23\)--\(120~\mathrm{ms}\) temporal resolution, and \(v_{\mathrm{enc}}=40\)--\(100~\mathrm{cm/s}\) for liver/kidney/brain and \(100\)--\(200~\mathrm{cm/s}\) for aorta/carotid.
Each case provides $N_e=4$ k-spaces of size \(N_t \times N_c \times N_x \times N_y \times N_z\), corresponding to its four image velocity encodings, alongside coil sensitivity maps, segmentation masks, and $v_{\mathrm{enc}}$.
The dimensions vary between cases.
Retrospective undersampling uses Gaussian \(k\)-\(t\) masks of size $N_t \times 1 \times N_x \times N_y \times 1$ at \(R \in [10,50]\), per case encoding.
In our experiments, for each four-point velocity encoded case \(\{\mathbf x_i\}_{i=0}^{3}\), the
\(R_i\times\)-accelerated k-space \(\mathbf y_i\) of encoding \(i\) is normalized as
\(\mathbf y_i\leftarrow
(\|\mathbf M_{R_i}\|_{\text{F}}/\|\mathbf y_i\|_2)\mathbf y_i\).

\paragraph{Training details.}
We train on 137 out of the 138 fully sampled aortic cases from the official training split and keep one for qualitative assessment. 
This is done on encoding patches of size \(P_t \times N_c \times N_x \times N_y \times P_z\), with $P_t = 15$, $P_z = 7$ and a batch size of 5. 
In a given batch, each element consists of a randomly selected encoding from the same four-point velocity-encoded case and is retrospectively undersampled at a random \(R \in [10,50]\).
We perform $50,000$ steps of the Adam optimizer with an initial learning rate of $5 \times 10^{-3}$ decaying to $1 \times 10^{-4}$ according to a cosine annealing schedule.
On an NVIDIA RTX PRO 6000 Blackwell, the training takes around $96$ hours.

\paragraph{Evaluation setup.}
We evaluate on all official validation sets, in which the cases are labeled as \(10\), \(20\), \(30\), \(40\) and \(50\times\)-accelerated. 
\textbf{Direct R1} evaluates the challenge's Regular Task~1 on 32 undersampled aortic cases, targeting accurate aorta reconstruction; 
\textbf{Generalizability S1} evaluates on the challenge's Special Task~1 on 40 cases from unseen sites and diseases;
\textbf{Generalizability S2} evaluates the challenge's Special Task~2 on 40 unseen-anatomy cases: 10 cerebrovascular, 10 portal vein, 10 renal artery, and 10 carotid cases.
Since validation ground truth is withheld, we report the official leaderboard metrics computed against hidden references within the segmentation mask: 
RelErr for vector-field magnitude error, AngErr for mean angular vector-field error, and nRMSE and SSIM for image magnitude, with SSIM averaged over the mask.
We compare against LLR, representing compressed sensing, and FlowVN, representing deep learning.
Both implementations are from the official challenge GitHub repository\footnote{\url{https://github.com/CmrxRecon/CMRx4DFlow2026} \label{challenge_repo}}, and FlowVN uses the organizer-provided trained weights.
The LLR-based reconstruction is computed using FISTA~\cite{beck2009fista}.
Comparisons against state-of-the-art methods are part of the official challenge leader-board and its final report.

\subsection{Quantitative and qualitative results}

% ============================================================
% Independent size controls
% ============================================================
\newcommand{\LeaderboardTableWidth}{1.00\textwidth}

\newcommand{\ComplexityPanelWidth}{0.43\textwidth}
\newcommand{\ComplexityTableWidth}{1.0\linewidth}

\newcommand{\AblationPanelWidth}{0.5\textwidth}
\newcommand{\AblationTableWidth}{1.0\linewidth}

\begin{table*}[!t]
\centering
\caption{
Quantitative performance, computational complexity, and acceleration-conditioning
ablation.
\textbf{(a)} Leaderboard results on Direct R1 and the generalizability sets S1
and S2.
\textbf{(b)} Model complexity and reconstruction time for a four-point-encoded, $30\times$-accelerated, 10-coil $k$-space case on a
$28 \times 84 \times 108 \times 18$ spatio-temporal grid, measured on an
NVIDIA RTX PRO 6000 Blackwell GPU.
\textbf{(c)} Comparison of acceleration-specific CLEAR models (at \(R=10,20,30,40,50\)) and the
acceleration-conditioned CLEAR model on Direct R1.
The best value in each comparison is shown in bold.
}
\label{tab:combined_results}

% ============================================================
% (a) Main leaderboard
% ============================================================
\textbf{(a) Leaderboard metrics}

\vspace{2pt}

\resizebox{\LeaderboardTableWidth}{!}{%
\begin{tabular}{@{}l*{12}{c}@{}}
\toprule
\multirow{2}{*}{\textbf{Method}}
& \multicolumn{4}{c}{\textbf{Direct R1}}
& \multicolumn{4}{c}{\textbf{Generalizability S1}}
& \multicolumn{4}{c}{\textbf{Generalizability S2}} \\
\cmidrule(lr){2-5}
\cmidrule(lr){6-9}
\cmidrule(lr){10-13}

& $\mathrm{RelErr}\!\downarrow$
& $\mathrm{AngErr}\!\downarrow$
& $\mathrm{nRMSE}\!\downarrow$
& $\mathrm{SSIM}\!\uparrow$

& $\mathrm{RelErr}\!\downarrow$
& $\mathrm{AngErr}\!\downarrow$
& $\mathrm{nRMSE}\!\downarrow$
& $\mathrm{SSIM}\!\uparrow$

& $\mathrm{RelErr}\!\downarrow$
& $\mathrm{AngErr}\!\downarrow$
& $\mathrm{nRMSE}\!\downarrow$
& $\mathrm{SSIM}\!\uparrow$ \\
\midrule

LLR
& 0.583 & 39.815 & 0.127 & 0.799
& 0.522 & 34.764 & 0.128 & 0.857
& 0.465 & 31.951 & 0.102 & 0.863 \\

FlowVN
& 0.526 & 41.562 & 0.059 & 0.927
& 0.626 & 40.599 & 0.110 & 0.909
& 0.803 & 44.044 & 0.120 & 0.854 \\

\textbf{CLEAR (ours)}
& \textbf{0.396}
& \textbf{30.214}
& \textbf{0.042}
& \textbf{0.959}

& \textbf{0.354}
& \textbf{23.211}
& \textbf{0.046}
& \textbf{0.983}

& \textbf{0.293}
& \textbf{24.800}
& \textbf{0.033}
& \textbf{0.989} \\

\bottomrule
\end{tabular}%
}

\vspace{7pt}

% ============================================================
% (b) Complexity and runtime
% (c) Conditioning ablation
% ============================================================
\begin{minipage}[t]{\ComplexityPanelWidth}
\vspace{0pt}
\centering

\textbf{(b) Complexity and reconstruction time}

\vspace{2pt}

\resizebox{\ComplexityTableWidth}{!}{%
\begin{tabular}{@{}lccc@{}}
\toprule
\textbf{Metric}
& \textbf{LLR}
& \textbf{FlowVN}
& \textbf{CLEAR (ours)} \\
\midrule

Recon. Time $\downarrow$
& 69 s
& \textbf{4 s}
& 64 s \\

\# of param. $\downarrow$
& --
& 64,860
& \textbf{8,317} \\

\bottomrule
\end{tabular}%
}

\end{minipage}%
\hfill
\begin{minipage}[t]{\AblationPanelWidth}
\vspace{0pt}
\centering

\textbf{(c) Acceleration-conditioning ablation}

\vspace{2pt}

\resizebox{\AblationTableWidth}{!}{%
\begin{tabular}{@{}lccccc@{}}
\toprule
\textbf{CLEAR variant}
& $\mathrm{RelErr}\!\downarrow$
& $\mathrm{AngErr}\!\downarrow$
& $\mathrm{nRMSE}\!\downarrow$
& $\mathrm{SSIM}\!\uparrow$
& $\mathrm{Train. Time}\!\downarrow$ \\
\midrule

Acceleration-specific
& 0.403
& 31.018
& 0.043
& 0.958
& 146 h \\

Acceleration-conditioned
& \textbf{0.396}
& \textbf{30.214}
& \textbf{0.042}
& \textbf{0.959}
& \textbf{96 h} \\

\bottomrule
\end{tabular}%
}

\end{minipage}

\end{table*}

Table~\ref{tab:combined_results} shows that CLEAR achieves the best performance across all validation settings and leaderboard metrics.
On Direct R1, it improves both velocity accuracy and magnitude reconstruction, reducing RelErr, AngErr, and nRMSE while increasing SSIM compared with LLR and FlowVN.
The gains are even larger on the generalizability sets S1 and S2, suggesting that the learned regularizer transfers well to unseen sites, diseases, and anatomical regions.
CLEAR is also parameter-efficient, using about eight times fewer parameters than FlowVN (\(8{,}317\) vs. \(64{,}860\)), while reconstruction time is comparable to that of LLR, though much slower than FlowVN.
Finally, the ablation shows that  acceleration-conditioned CLEAR  slightly outperforms the acceleration-specific trained with the same data, architecture, optimizer, and number of steps. Although the gain is small, conditioning is useful because the effective undersampling factors are not exactly the nominal values \(R=10,20,30,40,50\).
For instance, two cases labeled as \(30\times\) may correspond to true accelerations of \(30.22\times\) and \(30.07\times\).
The continuous parametrization in \eqref{eq:conditioning} adapts the potential scales to each such case within a single model, whereas acceleration-specific models use one fixed set of scales per nominal factor and require a separate training run for each.

\begin{figure}[!h]
    \centering

    \begin{minipage}[t]{0.49\textwidth}
        \centering
        \includegraphics[width=1.09\linewidth]{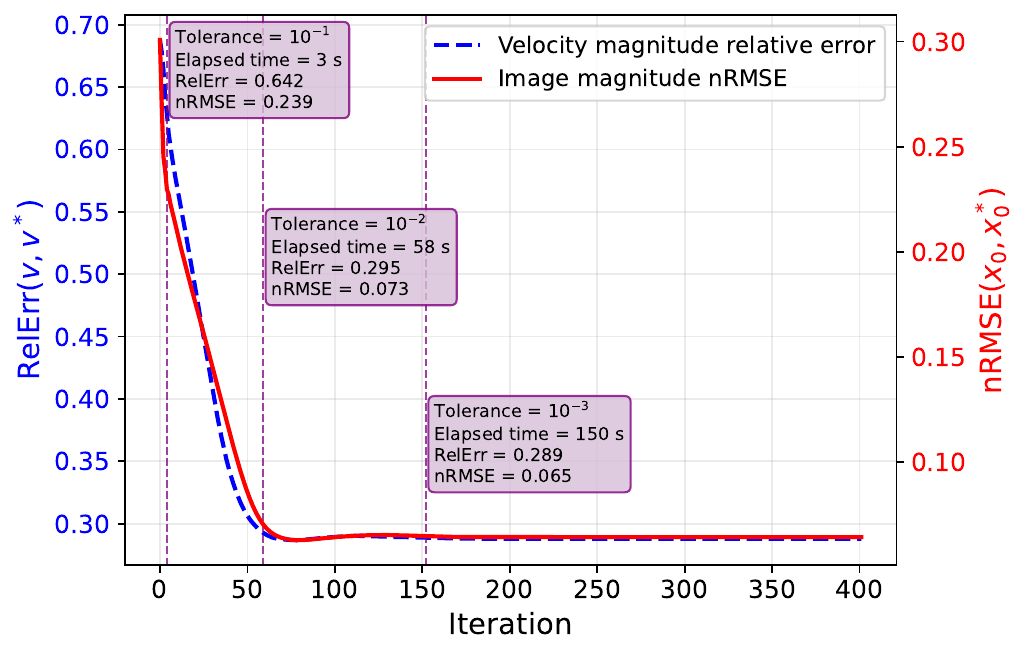}
        \textbf{(a)}
    \end{minipage}
    \hfill
    \begin{minipage}[t]{0.49\textwidth}
        \centering
        \includegraphics[width=0.91\linewidth]{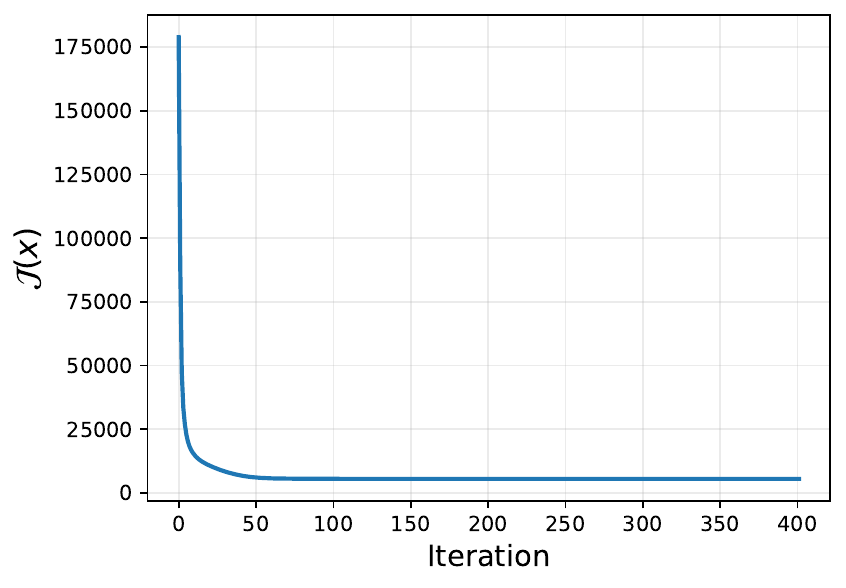}
        \textbf{(b)}
    \end{minipage}

    \caption{
    Convergence and stopping-tolerance sensitivity of CLEAR.
    (a) RelErr and nRMSE during batched reconstruction of the four velocity-encoded
    images in a 4D Flow CMR case, with markers indicating different nmAPG stopping
    tolerances. (b) Decrease of the objective \eqref{eq:optimization} during reconstruction of a
    single velocity-encoded image.
    }
    \label{fig:clear_convergence}
\end{figure}

To assess sensitivity to the stopping criterion of nmAPG, we also monitored reconstruction quality and objective decrease for different tolerances, as shown in Figure \ref{fig:clear_convergence}. Relaxing the tolerance from \(10^{-1}\) to \(10^{-2}\) improved the reconstruction (RelErr \(0.642\to0.295\), nRMSE \(0.239\to0.073\)), whereas tightening it to
\(10^{-3}\) gave only marginal gains (RelErr \(0.289\), nRMSE \(0.065\)) while
increasing runtime from \(58\) s to \(150\) s.
We therefore used an nmAPG tolerance of \(10^{-2}\) for all leaderboard evaluations, as it provided a favorable
accuracy--runtime compromise. The stable objective decrease further showcases optimization stability.

\begin{figure}[!t]
\includegraphics[width=\textwidth]{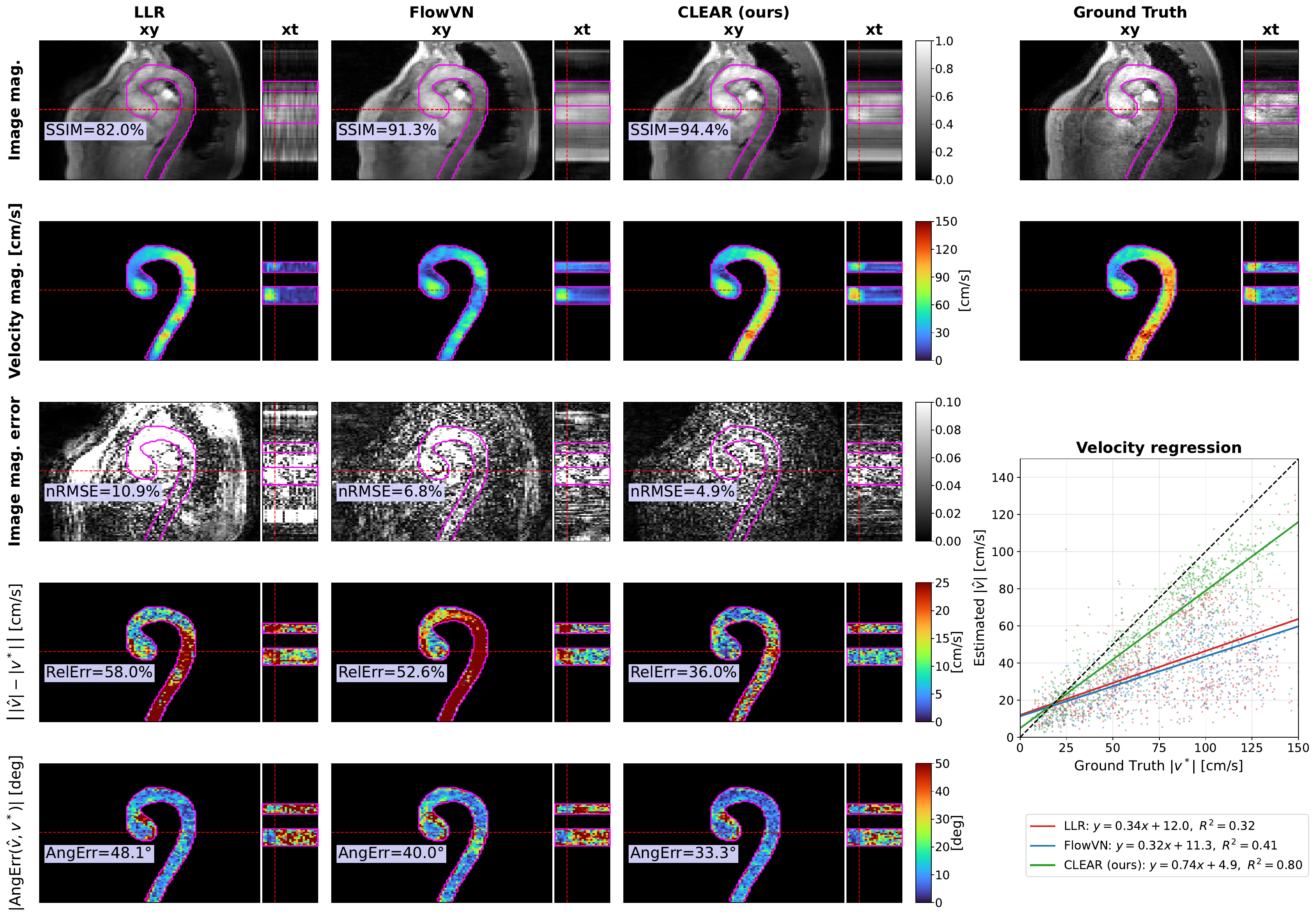}
\caption{Qualitative results on a  \(30\times\) undersampled aortic case. Corresponding slice locations are illustrated with red dashed lines, indicating
cross-section of the aorta and systolic peak. Scatter plot of velocity magnitude over manually segmented aorta (contour shown in magenta) is given together with correlation analysis ($y = ax + b$).} \label{fig:qualitative_assessment}
\end{figure}

Figure~\ref{fig:qualitative_assessment} shows qualitative results on a representative \(30\times\) undersampled aortic case.
Compared with LLR and FlowVN, CLEAR better preserves the magnitude image and substantially reduces both magnitude and velocity errors within the segmented aorta. This is reflected by the improved image metrics
(SSIM \(=94.4\%\), nRMSE \(=4.9\%\)) and velocity metrics
(RelErr \(=36.0\%\), AngErr \(=33.5^\circ\)). The velocity scatter plot further shows that CLEAR provides the closest agreement with the reference velocities,
with a slope closer to the identity and a markedly higher correlation (\(R^2=0.80\)) than LLR (\(R^2=0.32\)) and FlowVN (\(R^2=0.41\)). 
Although high velocities remain slightly underestimated, CLEAR produces the least biased and most spatially consistent reconstruction among the compared methods.

\subsection{Interpretability} \label{sec:interpretability}

\begin{figure}[!t]
\includegraphics[width=\textwidth]{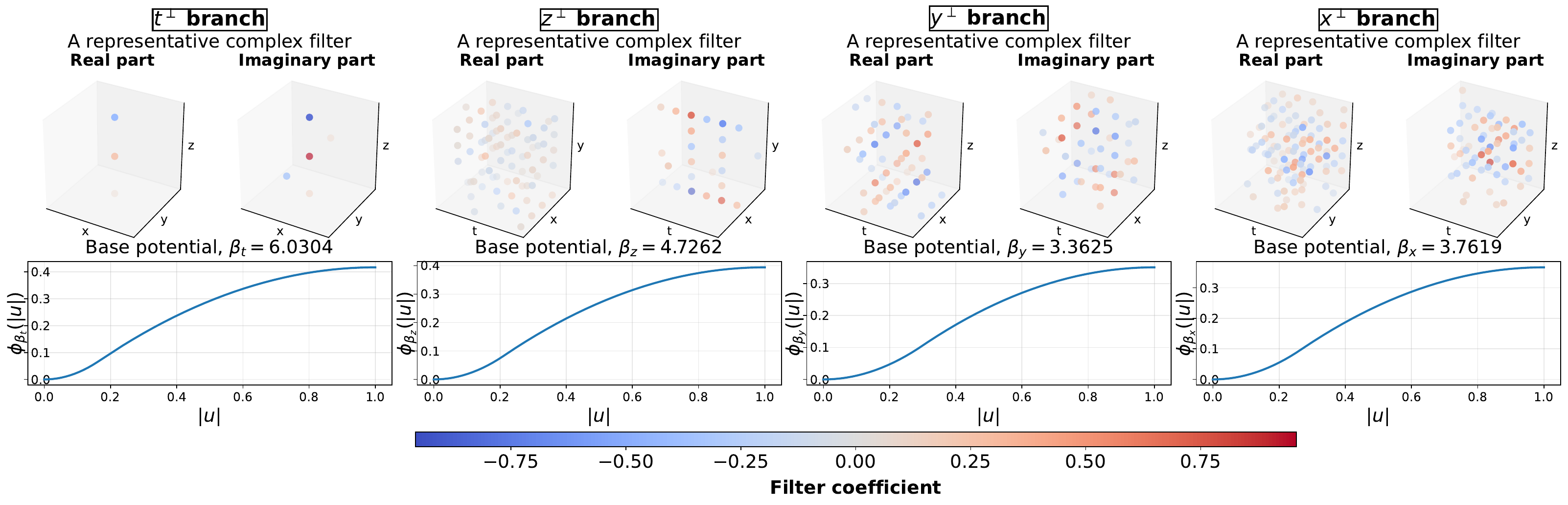}
\caption{Visualization of the learned components in CLEAR.
Only filter coefficients with magnitudes larger than \(15\%\) of the maximum magnitude are displayed for clarity.} \label{fig:learned_components}
\end{figure}

Figure~\ref{fig:learned_components} shows representative filters and potentials
from the four 3D regularizer branches \eqref{eq:regularizer_3d_branch} in CLEAR. The \(t^\perp\) branch learns a spatial high-pass
filter, visually reminiscent of finite differences or local contrast detectors, whereas the \(z^\perp\), \(y^\perp\), and \(x^\perp\) branches learn broader oscillatory
spatio-temporal filters resembling higher-order derivatives and wavelet-like patterns. The learned potentials act on the magnitudes of the complex filter responses. Their smooth, saturating shape penalizes small responses strongly while treating large responses more mildly, encouraging incoherent features to vanish while preserving strong anatomical and flow-related structures. Thus, the filters indicate which local features are measured, and the potentials indicate how sparsity is imposed on them.

\section{Discussion and Conclusion}

We introduced CLEAR, a complex, learned, explicit regularizer for 4D Flow CMR reconstruction.
The method combines multi-subspace 3D complex filters, smooth sparsity-promoting potentials, and acceleration conditioning within a variational formulation.
Across the CMRx4DFlow2026 validation settings, CLEAR consistently improved magnitude and velocity metrics over the baselines with less than 10k parameters.
The results suggest that learned regularizers can bridge part of the gap between classical compressed sensing and deep reconstruction networks.
This is particularly relevant for clinical reconstruction, where robust and interpretable reconstruction methods are desirable.

The current design, however, has two limitations.
First, the regularizer uses a smooth magnitude \(|\cdot|_{\varepsilon_0}\) to map each complex filter response to a real scalar before applying the potential.
This radial construction is simple and smooth but invariant to the phase of the complex response and thus may discard useful phase-dependent information in the learned feature space.
Future work could therefore investigate more expressive and phase-aware mappings.
Second, the regularizer processes each image encoding within a four-point velocity-encoded 4D Flow CMR case independently, which leads to independent reconstructions of each image encoding.
A joint processing of the four encodings within a case, as done in \cite{jacobs2025flowmrinet}, would exploit the correlations between them and potentially lead to improved reconstructions.
This is also left for future work, as well as prospective validation, uncertainty estimation, and clinical reader studies.

\subsubsection{Code availability.}  The source code and trained weights for CLEAR are publicly available on GitHub: \url{https://github.com/Shamachrist7/CLEAR}.

\subsubsection{\ackname} This study was funded by the DFG within the SPP2298 under the Project Number 543939932.

% \subsubsection{\discintname}
% The authors have no competing interests to declare that are
% relevant to the content of this article.

%
% ---- Bibliography ----
%
% BibTeX users should specify bibliography style 'splncs04'.
% References will then be sorted and formatted in the correct style.
%
% \bibliographystyle{splncs04}
% \bibliography{mybibliography}

\begin{thebibliography}{8}

\bibitem{markl2012time}
Markl, M., Frydrychowicz, A., Kozerke, S., Hope, M., Wieben, O.:
4D flow MRI. J. Magn. Reson. Imaging 36(5), 1015--1036 (2012).
\doi{10.1002/jmri.23632}

\bibitem{dyverfeldt2015}
Dyverfeldt, P., Bissell, M., Barker, A.J., et al.:
4D flow cardiovascular magnetic resonance consensus statement.
J. Cardiovasc. Magn. Reson. 17, 72 (2015).
\doi{10.1186/s12968-015-0174-5}

\bibitem{bissell2023consensus}
Bissell, M.M., Raimondi, F., Ait Ali, L., et al.:
4D Flow cardiovascular magnetic resonance consensus statement: 2023 update.
Journal of Cardiovascular Magnetic Resonance 25, 40 (2023).
\doi{10.1186/s12968-023-00942-z}

\bibitem{rudin1992tv}
Rudin, L.I., Osher, S., Fatemi, E.:
Nonlinear total variation based noise removal algorithms.
Physica D 60(1--4), 259--268 (1992).
\doi{10.1016/0167-2789(92)90242-F}

\bibitem{trzasko2011llr}
Trzasko, J., Manduca, A.:
Local versus global low-rank promotion in dynamic MRI series reconstruction.
In: Proc. ISMRM, p. 4371 (2011)

\bibitem{lyu2023regionfocused}
Lyu, J., Li, G., Wang, C., et al.:
Region-focused multi-view transformer GAN for cardiac cine MRI reconstruction.
Med. Image Anal. 85, 102760 (2023).
\doi{10.1016/j.media.2023.102760}

\bibitem{lyu2023federated}
Lyu, J., Tian, Y., Cai, Q., et al.:
Adaptive channel-modulated personalized federated learning for MRI reconstruction.
Comput. Biol. Med. 165, 107330 (2023).
\doi{10.1016/j.compbiomed.2023.107330}

\bibitem{wang2023odls}
Wang, Z., Qian, C., Guo, D., et al.:
One-dimensional deep low-rank and sparse network for accelerated MRI.
IEEE Trans. Med. Imaging 42, 79--90 (2023).
\doi{10.1109/TMI.2022.3203312}

\bibitem{qin2019crnn}
Qin, C., Schlemper, J., Caballero, J., et al.:
Convolutional recurrent neural networks for dynamic MR image reconstruction.
IEEE Trans. Med. Imaging 38, 280--290 (2019).
\doi{10.1109/TMI.2018.2863670}

\bibitem{lyu2024stadnet}
Lyu, J., Wang, S., Tian, Y., et al.:
STADNet: spatial-temporal attention-guided dual-path network for cardiac cine MRI super-resolution.
Med. Image Anal. 94, 103142 (2024).
\doi{10.1016/j.media.2024.103142}

\bibitem{wang2025deepseparable}
Wang, Z., Xiao, M., Zhou, Y., et al.:
Deep separable spatiotemporal learning for fast dynamic cardiac MRI.
IEEE Trans. Biomed. Eng. 72, 3642--3654 (2025).
\doi{10.1109/TBME.2025.3574090}

\bibitem{huang2025mambamir}
Huang, J., Yang, L., Wang, F., et al.:
Enhancing global sensitivity and uncertainty quantification in medical image reconstruction with Monte Carlo arbitrary-masked Mamba.
Med. Image Anal. 99, 103334 (2025).
\doi{10.1016/j.media.2024.103334}

\bibitem{wang2025pisf}
Wang, Z., Yu, X., Wang, C., et al.:
One for multiple: physics-informed synthetic data boosts generalizable deep learning for fast MRI reconstruction.
Med. Image Anal. 103, 103616 (2025).
\doi{10.1016/j.media.2025.103616}

\bibitem{lyu2026dpinr}
Lyu, J., Wang, G., Wang, Z., et al.:
Diffusion-prior implicit neural representation for arbitrary-scale cardiac cine MRI super-resolution.
Inf. Fusion 126, 103510 (2026).
\doi{10.1016/j.inffus.2025.103510}

\bibitem{vishnevskiy2020flowvn}
Vishnevskiy, V., Walheim, J., Kozerke, S.:
Deep variational network for rapid 4D flow MRI reconstruction.
Nat. Mach. Intell. 2, 228--235 (2020).
\doi{10.1038/s42256-020-0165-6}

\bibitem{jacobs2025flowmrinet}
Jacobs, L., Piccirelli, M., Vishnevskiy, V., Kozerke, S.:
FlowMRI-Net: a generalizable self-supervised 4D flow MRI reconstruction network.
J. Cardiovasc. Magn. Reson. 27, 101913 (2025).
\doi{10.1016/j.jocmr.2025.101913}

\bibitem{roth2009fields}
Roth, S., Black, M.J.:
Fields of experts. Int. J. Comput. Vis. 82(2), 205--229 (2009).
\doi{10.1007/s11263-008-0197-6}

\bibitem{goujon2023convex}
Goujon, A., Neumayer, S., Bohra, P., Ducotterd, S., Unser, M.:
A neural-network-based convex regularizer for inverse problems.
IEEE Trans. Comput. Imaging 9, 781--795 (2023).
\doi{10.1109/TCI.2023.3306100}

\bibitem{goujon2024weakly}
Goujon, A., Neumayer, S., Unser, M.:
Learning weakly convex regularizers for convergent image-reconstruction algorithms.
SIAM J. Imaging Sci. 17(1), 91--115 (2024).
\doi{10.1137/23M1565243}

\bibitem{wache2026wcrr}
Wache, G.S., Chaithya, G.R., Tanabene, A., Neumayer, S.:
Weakly convex ridge regularization for 3D non-Cartesian MRI reconstruction.
arXiv:2603.27158 (2026)

\bibitem{hertrich2026learning}
Hertrich, J., Wong, H.S., Denker, A., et al.:
Learning regularization functionals for inverse problems: a comparative study.
Handbook of Numerical Analysis, Elsevier (2026).
\doi{10.1016/bs.hna.2026.04.001}

\bibitem{bolte2023onestep}
Bolte, J., Pauwels, E., Vaiter, S.:
One-step differentiation of iterative algorithms.
In: NeurIPS, vol. 36, pp. 77089--77103 (2023)

\bibitem{fung2022jfb}
Fung, S.W., Heaton, H., Li, Q., McKenzie, D., Osher, S., Yin, W.:
JFB: Jacobian-free backpropagation for implicit networks.
Proc. AAAI 36(6), 6648--6656 (2022).
\doi{10.1609/aaai.v36i6.20619}

\bibitem{li2015apg}
Li, H., Lin, Z.:
Accelerated proximal gradient methods for nonconvex programming.
In: NeurIPS, vol. 28, pp. 379--387 (2015)

\bibitem{wang2024cmrxrecon}
Wang, C., Lyu, J., Wang, S., et al.:
CMRxRecon: a publicly available k-space dataset and benchmark for cardiac MRI.
Sci. Data 11, 687 (2024).
\doi{10.1038/s41597-024-03525-4}

\bibitem{wang2025cmrxrecon2024}
Wang, Z., Wang, F., Qin, C., et al.:
CMRxRecon2024: a multimodality, multiview k-space dataset for accelerated cardiac MRI.
Radiol. Artif. Intell. 7, e240443 (2025).
\doi{10.1148/ryai.240443}

\bibitem{wang2025foundation}
Wang, Z., Huang, M., Shi, Z., et al.:
Enabling ultra-fast cardiovascular imaging across heterogeneous clinical environments with a generalist foundation model and multimodal database.
arXiv:2512.21652 (2025)

\bibitem{lyu2025cmrxreconchallenge}
Lyu, J., Qin, C., Wang, S., et al.:
State-of-the-art cardiac MRI reconstruction: results of the CMRxRecon challenge.
Med. Image Anal. 101, 103485 (2025).
\doi{10.1016/j.media.2025.103485}

\bibitem{wang2025cmrxmotion}
Wang, K., Qin, C., Shi, Z., et al.:
Extreme cardiac MRI analysis under respiratory motion: results of the CMRxMotion challenge.
Med. Image Anal., 103883 (2025).
\doi{10.1016/j.media.2025.103883}

\bibitem{wang2026cmrxrecon2024summary}
Wang, F., Wang, Z., Li, Y., et al.:
Toward modality- and sampling-universal learning strategies for accelerating cardiovascular imaging.
IEEE Trans. Med. Imaging 45, 1872--1887 (2026).
\doi{10.1109/TMI.2025.3641610}

\bibitem{beck2009fista}
Beck, A., Teboulle, M.:
A fast iterative shrinkage-thresholding algorithm for linear inverse problems.
SIAM Journal on Imaging Sciences 2(1), 183--202 (2009).
\doi{10.1137/080716542}

\bibitem{wang2021phenomics}
Wang, C., Li, Y., Lv, J., et al.:
Recommendation for CMR imaging-based phenotypic study: imaging part.
Phenomics 1, 151--170 (2021).
\doi{10.1007/s43657-021-00018-x}


\end{thebibliography}
%

\renewcommand{\thesection}{S}
\renewcommand{\theequation}{S\arabic{equation}}
\renewcommand{\thefigure}{S\arabic{figure}}
\renewcommand{\thetable}{S\arabic{table}}

\section*{Supplementary material}

\begin{figure}[!h]
\includegraphics[width=\textwidth]{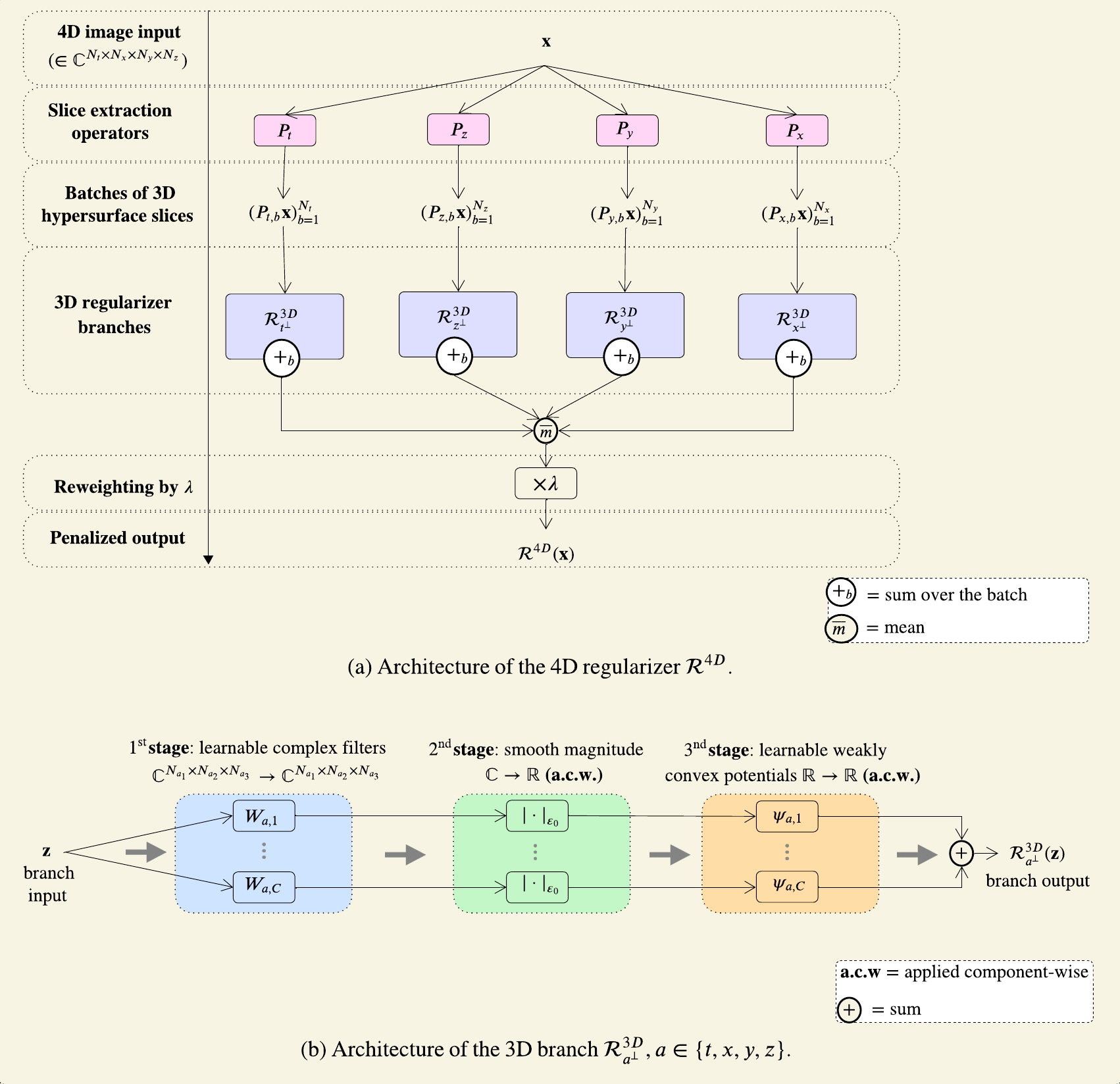}
\caption{Illustration of the proposed regularizer architecture.} \label{fig:reg}
\end{figure}

\subsection{Jacobian-free backpropagation training details}
Let \(\theta\) be the collection of all learnable parameters, and write
\(\mathcal R^{4D}_{\theta(R)}\) to make the dependence on the acceleration factor \(R\) explicit.
An $R\times$-accelerated k-space $\mathbf y \in \mathbb{C}^{N_t \times N_c \times N_x \times N_y \times N_z}$ is reconstructed as $\widehat{\mathbf x} \in \mathbb{C}^{N_t \times N_x \times N_y \times N_z}$ according to 
\begin{equation} \label{eq:var_recon}
    \widehat{\mathbf x}(\mathbf y, \theta)
    =
    \arg\min_{\mathbf x}
    \frac12\|\mathbf{M}_R\mathcal{F}\mathbf{S} \mathbf x-\mathbf y\|_2^2
    +
    \mathcal R^{4D}_{\theta(R)}(\mathbf x),
\end{equation}
which we solve using the nmAPG solver.

At the Adam optimizer step \(j\), let
\(\{(\mathbf x^{j,b},\mathbf y^{j,b})\}_{b\in\mathcal B}\) be the training batch, where \(\mathbf y^{j,b}\) is $R^{j,b}\times$-accelerated.
A direct parameter update would require backpropagating through
\(
    \sum_{b\in\mathcal B}
    \|
        \widehat{\mathbf x}(\mathbf y^{j,b},\theta)
        -
        \mathbf x^{j,b}
    \|_2^2,
\)
where \(\widehat{\mathbf x}(\mathbf y^{j,b},\theta)\) is obtained by solving the lower-level variational problem \eqref{eq:var_recon} with nmAPG.
Instead, we detach the nmAPG reconstruction
\(\widehat{\mathbf x}(\mathbf y^{j,b},\theta)\) from the computational graph and take one additional gradient step:
\begin{equation} \label{eq:last_step}
    \widetilde{\mathbf x}(y^{j,b},\theta)
    =
    \widehat{\mathbf x}(y^{j,b},\theta)
    -
    \frac{1}{L_{j,b}}
    \nabla_{\mathbf x}
    \left[
    \frac12\|\mathbf{M}_{R^{j,b}}\mathcal{F}\mathbf{S} \mathbf x-\mathbf y^{j,b}\|_2^2
    +
    \mathcal R^{4D}_{\theta(R_{j,b})}(\mathbf x)
    \right]_{\mathbf x=\widehat{\mathbf x}(y^{j,b},\theta)}.   
\end{equation}

Here, \(L_{j,b}\) is the local Lipschitz of the variational objective. The parameters \(\theta\) are then updated by backpropagating only through this last step, using
\(
    \sum_{b\in\mathcal B}
    \|
        \widetilde{\mathbf x}(\mathbf y^{j,b},\theta)
        -
        \mathbf x^{j,b}
    \|_2^2.
\)

\subsection{Reported metrics}

Let \(\mathbf x\) and \(\mathbf x^\ast\) denote the reconstructed and
reference magnitude images of the four-point velocity-encoding, and let
\(\widehat{\mathbf v}\) and \(\mathbf v^\ast\) denote the reconstructed and
reference velocity fields. We reported
\[
\begin{aligned}
\mathrm{nRMSE}(\mathbf x,\mathbf x^\ast)
&=
\sqrt{
\sum_i
\frac{
    (x_{i}-x_{i}^\ast)^2
}{
    N\max_j (x_{j}^\ast)^2
}
}, \\[2mm]
\mathrm{RelErr}(\widehat{\mathbf v},\mathbf v^\ast)
&=
\frac{
    \big\|
        |\widehat{\mathbf v}|-|\mathbf v^\ast|
    \big\|_2
}{
    \big\|
        |\mathbf v^\ast|
    \big\|_2
}, \\[2mm]
\mathrm{AngErr}(\widehat{\mathbf v},\mathbf v^\ast)
&=
\arccos
\left(
\frac{
    \langle \widehat{\mathbf v},\mathbf v^\ast\rangle
}{
    \|\widehat{\mathbf v}\|_2\|\mathbf v^\ast\|_2
}
\right).
\end{aligned}
\]
Additionally, we reported the structural similarity index (SSIM) on the reconstructed four-point velocity-encoding magnitude images.

\subsection{Tuning details for LLR}

The LLR baseline promotes low-rank structure in local patch matrices. Given a 4D image \(\mathbf x\), the LLR penalty is
\[
    \mathcal R_{\mathrm{LLR}}(\mathbf x)
    =
    \lambda_{\mathrm{LLR}}
    \sum_{i=1}^{N_{\mathrm{ptch}}}
    \|\mathbf T_i\mathbf x\|_* ,
\]
where \(\mathbf T_i\) extracts the \(i\)-th local \(p\times p\times p\) patch from all $N_t$ cardiac frames, \(N_{\mathrm{ptch}}\) is the number of patches, and \(\|\cdot\|_*\) denotes the nuclear norm. The corresponding convex optimization problem is solved with FISTA and stopped when the relative change between consecutive iterates falls below \(10^{-3}\), after which the RelErr remains unchanged. We tuned \(\lambda_{\mathrm{LLR}}\) separately for each acceleration factor present in the evaluation data ($R = 10, 20, 30, 40, 50$). This was done on the training data by grid search to minimize RelErr, obtaining \(\lambda_{\mathrm{LLR}}=0.4,0.6,0.6,0.8,0.8\) for
\(R=10,20,30,40,50\), respectively.

\subsection{CLEAR's learned components (full visualization)}

In each 3D branch, the per-filter potentials are just a rescaling of the base potential and keep the same shape. We therefore only present the base potential in each branch.

\begin{figure}[!h]
    \centering
    \includegraphics[page=1,width=\linewidth]{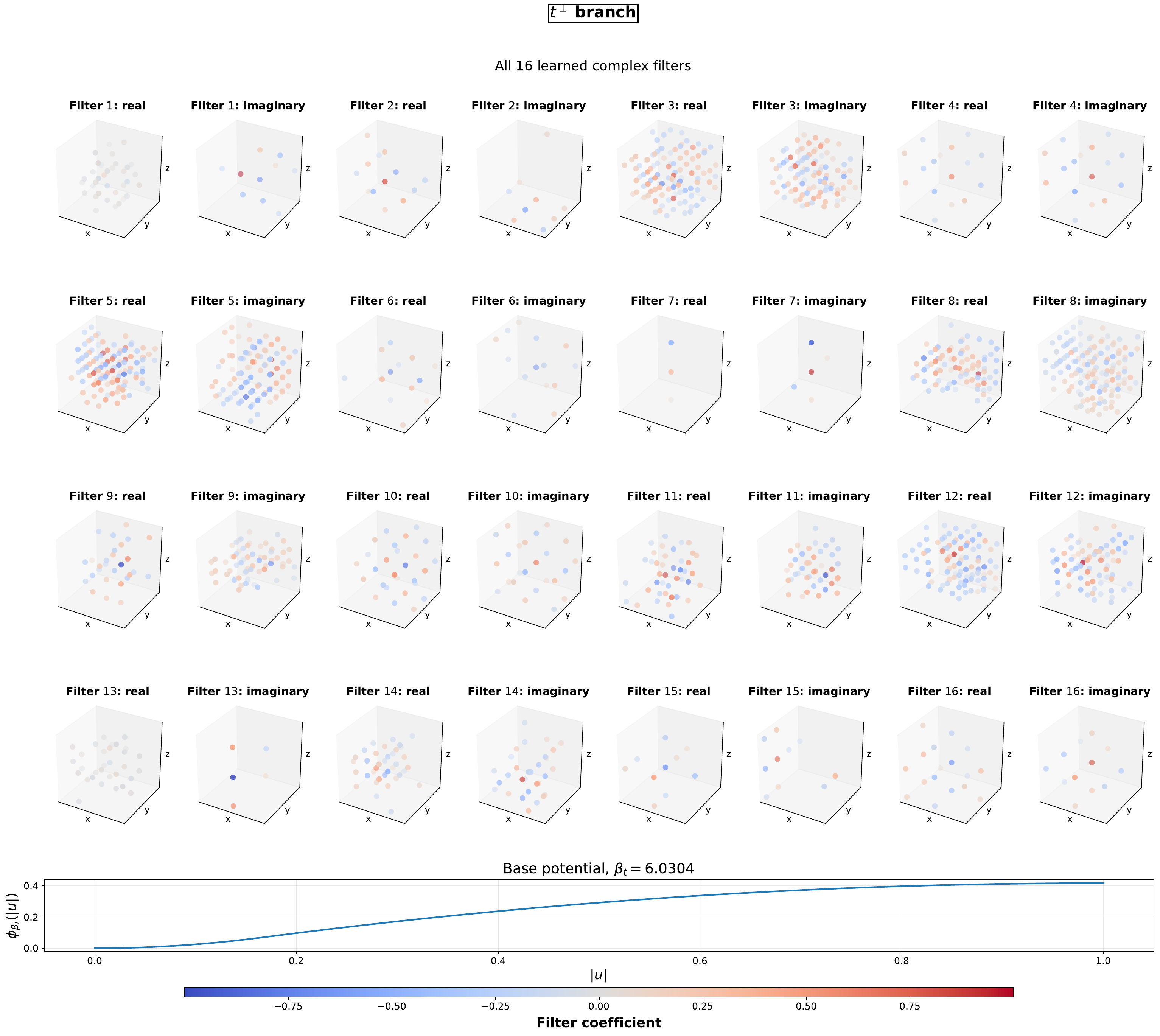}
    \caption{CLEAR's learned components (complex filters and base potential) in the $t^\perp$ 3D branch.  Only filter coefficients with magnitudes larger than \(15\%\) of the maximum magnitude filter coefficient are displayed for clarity.}
    \label{}
\end{figure}

\begin{figure}[!h]
    \centering
    \includegraphics[page=2,width=\linewidth]{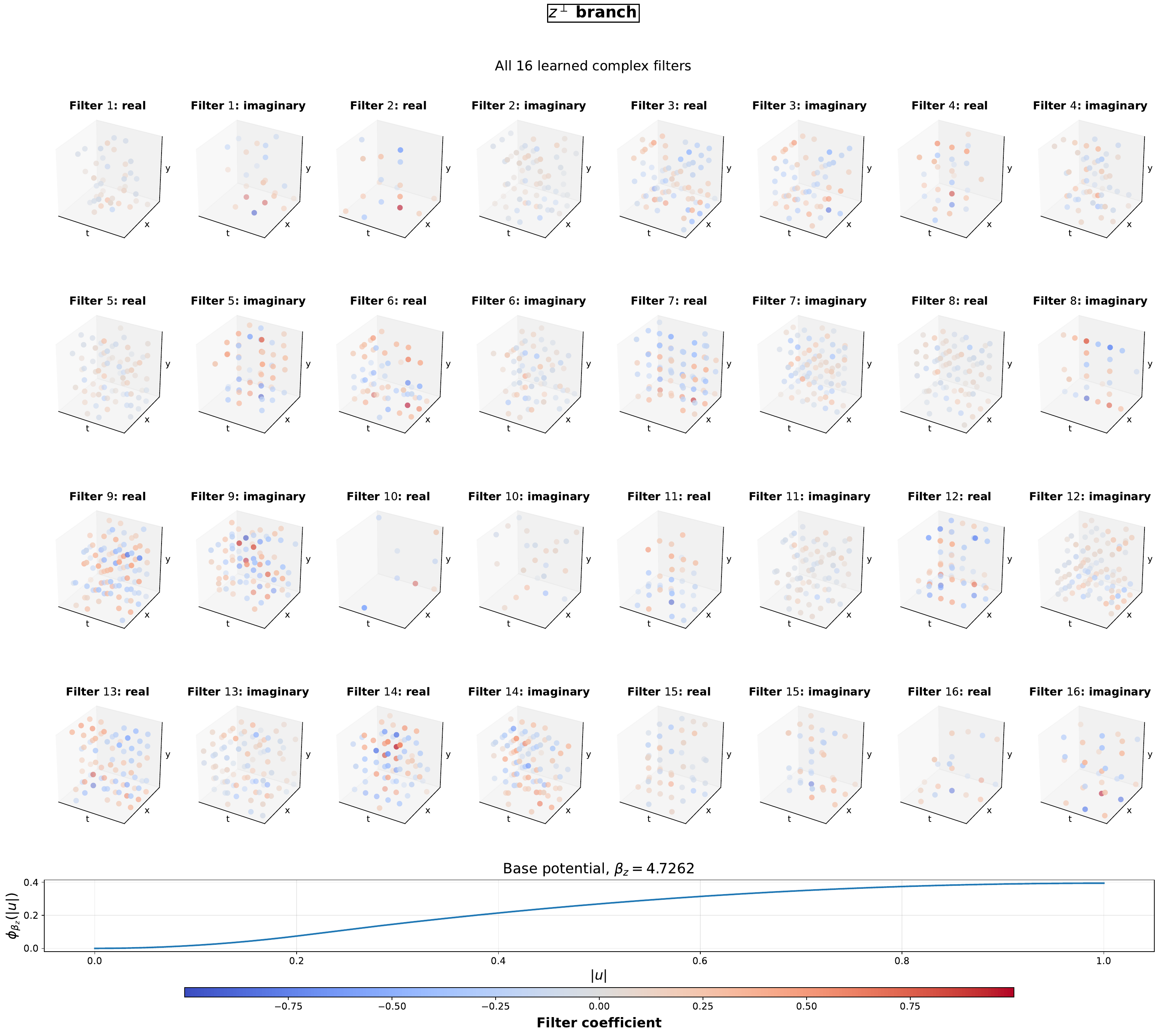}
    \caption{CLEAR's learned components (complex filters and base potential) in the $z^\perp$ 3D branch.  Only filter coefficients with magnitudes larger than \(15\%\) of the maximum magnitude filter coefficient are displayed for clarity.}
    \label{}
\end{figure}

\begin{figure}[!h]
    \centering
    \includegraphics[page=3,width=\linewidth]{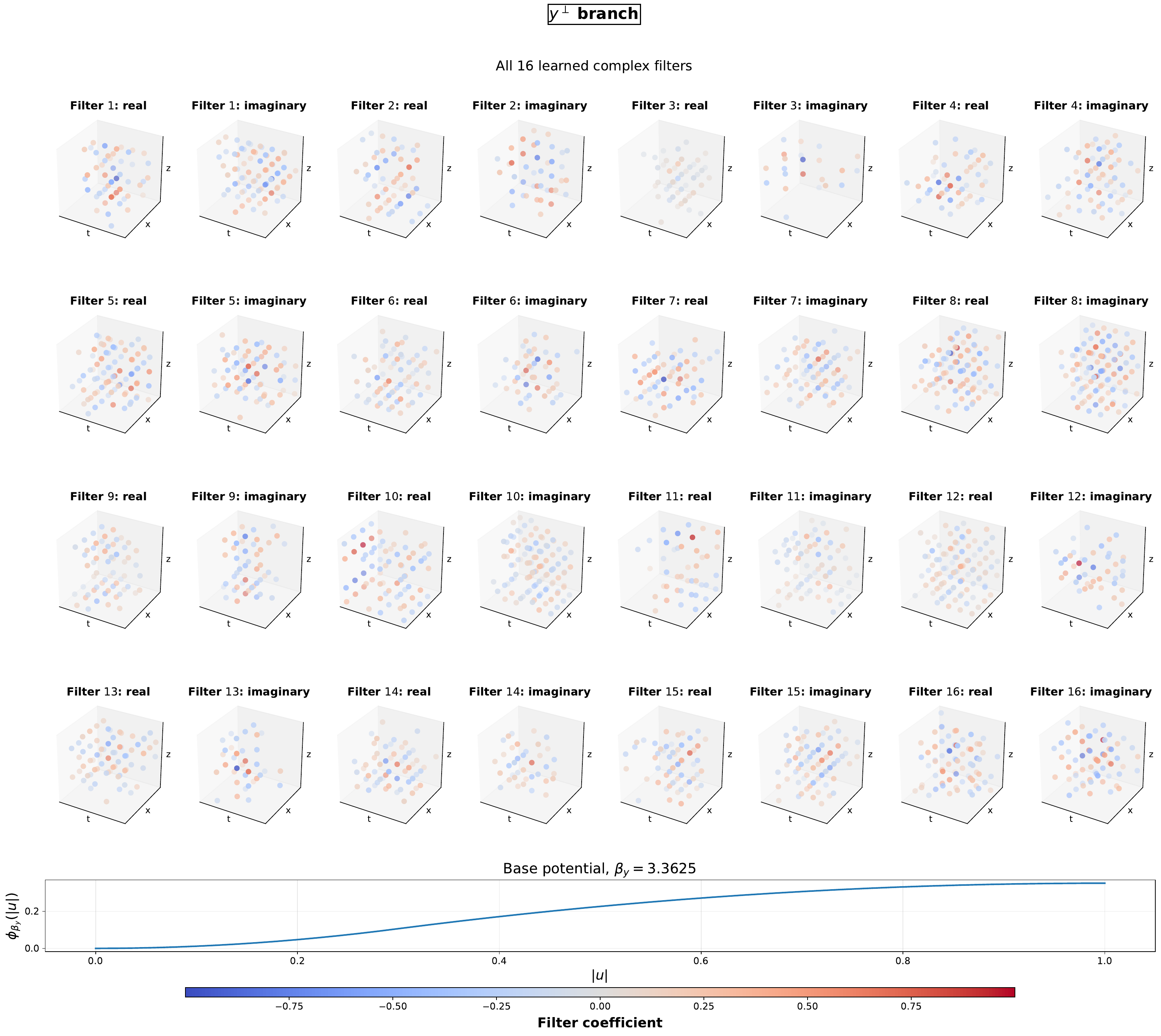}
    \caption{CLEAR's learned components (complex filters and base potential) in the $y^\perp$ 3D branch.  Only filter coefficients with magnitudes larger than \(15\%\) of the maximum magnitude filter coefficient are displayed for clarity.}
    \label{}
\end{figure}

\begin{figure}[!h]
    \centering
    \includegraphics[page=4,width=\linewidth]{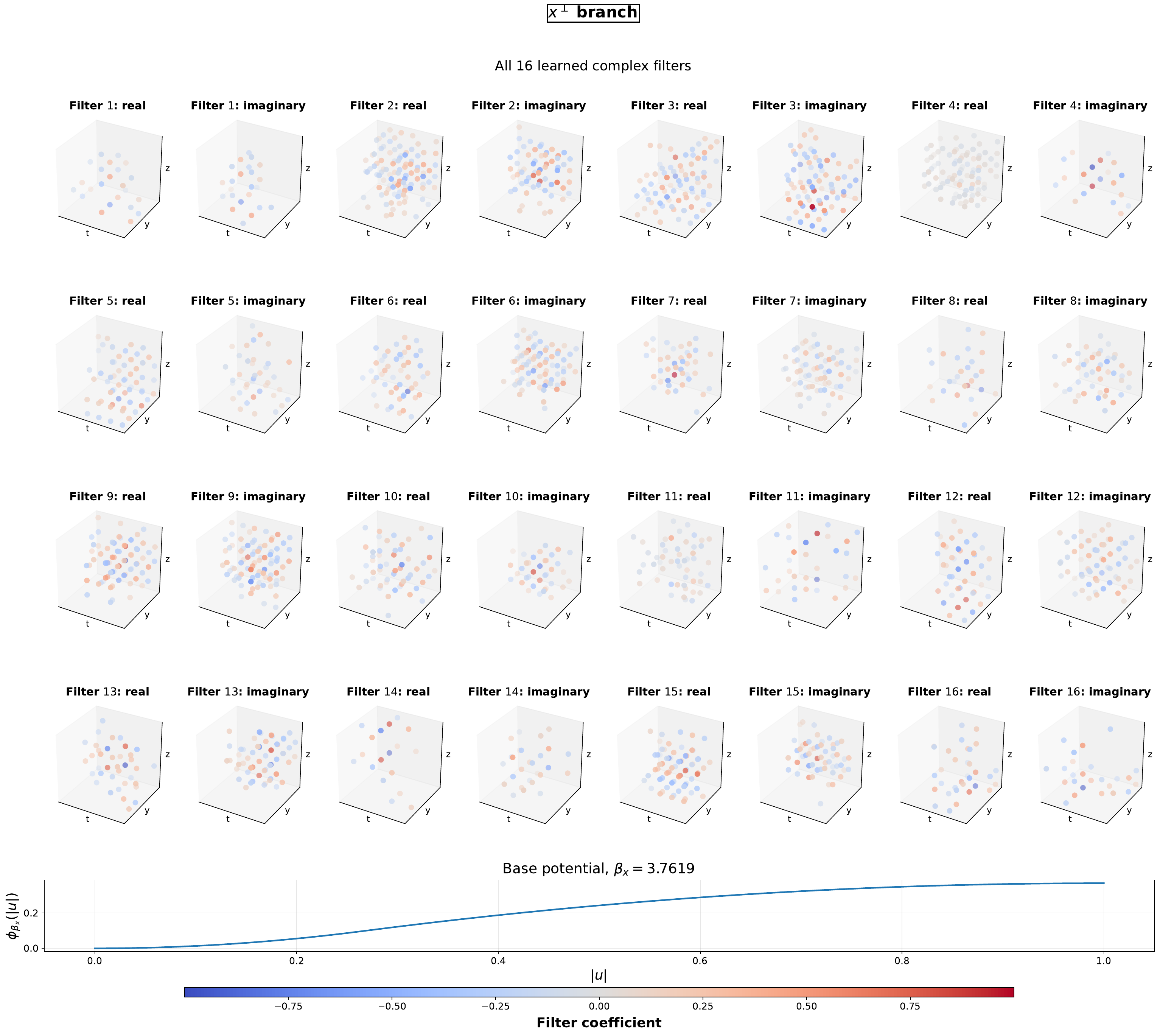}
    \caption{CLEAR's learned components (complex filters and base potential) in the $x^\perp$ 3D branch.  Only filter coefficients with magnitudes larger than \(15\%\) of the maximum magnitude filter coefficient are displayed for clarity.}
    \label{}
\end{figure}

\end{document}